\documentclass{article}
\usepackage{ijcai24}

\usepackage{times}
\usepackage{soul}
\usepackage{url}
\usepackage[hidelinks]{hyperref}
\usepackage[utf8]{inputenc}
\usepackage[small]{caption}
\usepackage{graphicx}
\usepackage{amsmath}
\usepackage{amsthm}
\usepackage{booktabs}
\usepackage{algorithm}
\usepackage{algorithmic}
\usepackage[switch]{lineno}
\usepackage{multirow}

\title{PPTFormer: Pseudo Multi-Perspective Transformer for UAV Segmentation}

\author{
Deyi Ji$^{1,2}$ \and
Wenwei Jin$^{2}$ \and
Hongtao Lu$^3$\And
Feng Zhao$^1$\thanks{Corresponding author.} \\
\affiliations
$^1$University of Science and Technology of China \\
$^2$Alibaba Group\\
$^3$Dept. of CSE, MOE Key Lab of Artificial Intelligence, AI Institute, Shanghai Jiao Tong University \\
jideyi@mail.ustc.edu.cn, fzhao956@ustc.edu.cn
}

\begin{document}

\maketitle

\begin{abstract}
    The ascension of Unmanned Aerial Vehicles (UAVs) in various fields necessitates effective UAV image segmentation, which faces challenges due to the dynamic perspectives of UAV-captured images. Traditional segmentation algorithms falter as they cannot accurately mimic the complexity of UAV perspectives, and the cost of obtaining multi-perspective labeled datasets is prohibitive. To address these issues, we introduce the PPTFormer, a novel \textbf{P}seudo Multi-\textbf{P}erspective \textbf{T}rans\textbf{former} network that revolutionizes UAV image segmentation. Our approach circumvents the need for actual multi-perspective data by creating pseudo perspectives for enhanced multi-perspective learning. The PPTFormer network boasts Perspective Representation, novel Perspective Prototypes, and a specialized encoder and decoder that together achieve superior segmentation results through Pseudo Multi-Perspective Attention  (PMP Attention) and fusion. Our experiments demonstrate that PPTFormer achieves state-of-the-art performance across five UAV segmentation datasets, confirming its capability to effectively simulate UAV flight perspectives and significantly advance segmentation precision. This work presents a pioneering leap in UAV scene understanding and sets a new benchmark for future developments in semantic segmentation.

\end{abstract}

\section{Introduction}
\label{sec:intro}

Semantic segmentation is a fundamental component of many practical applications in computer vision \cite{chen2023sam,yang2024irvr,yang2022unleashing,yang2022sir,yang2023visual,chen2024xlstm,yu2022deep,yu2024deep}, such as autonomous driving, video surveillance, and precision agriculture, where the goal is to predict a category label for each pixel in an image. In recent years, a plethora of algorithms and networks \cite{stlnet,zhu2023add,urur,sstkd,zhu2023gabor,zhu2023continual} have been developed for segmentation in conventional scenes, showcasing significant advancements in the field.

With the burgeoning growth of Unmanned Aerial Vehicles (UAVs), UAV segmentation has emerged as a pivotal area of research, playing a crucial role in applications ranging from environmental monitoring to disaster response. Unlike data captured from fixed perspectives, UAVs operate at varying altitudes and angles, offering a wealth of dynamic viewpoints.
The images captured by UAVs reflect the rich and varied visual information from different angles and altitudes, which is of great importance for monitoring rapidly changing environments.

Addressing the segmentation challenges posed by such rich variability in perspectives necessitates a novel approach. The most intuitive solution would be to collect actual multi-perspective data for training segmentation networks. However, as analyzed in \cite{dlpl,farseg,sco,homview}, due to the prohibitive costs of collection and fine-grained annotation, existing datasets lack multi-perspective images with detailed labeling. Conventional segmentation methods typically rely on explicit perspective transformation-based data augmentation techniques, such as scaling, rotating, or flipping images in 2D or 3D dimensions. These rudimentary methods of augmentation produce stiff and unnatural perspectives that fail to represent the true changes in perspective experienced during UAV flight, resulting in limited model performance in real-world UAV scenarios.

In light of these challenges and the rapid development of Vision Transformers in semantic segmentation, a subset of methodologies \cite{segformer,setr} has been applied to UAV scene segmentation. However, these Transformer networks have not deeply analyzed or been designed with UAV perspective dynamics in mind. The DLPL \cite{dlpl} introduces a comprehensive approach for latent perspective learning. However, DLPL employs a more intricate method involving discrete perspective decomposition and statistical representation, and relies on Homography transformations for converting latent perspectives. Consequently, this framework requires a slightly larger number of parameters and incurs higher latency. In this paper, we adopt the theoretical foundation of DLPL to propose a more lightweight and innovative network, termed PPTFormer (\textbf{P}seudo Multi-\textbf{P}erspective \textbf{T}rans\textbf{former}), specifically tailored for UAV segmentation tasks. The core of this network is composed of advanced PPTFormer blocks designed for pseudo multi-perspective learning. These blocks utilize perspective prototypes that remain consistent throughout the network, thereby enhancing perspective-aware learning. Drawing inspiration from \cite{dlpl}, a critical component of the PPTFormer Blocks is the perspective representation and generation module. This module modifies visual features to emulate various UAV viewpoints while maintaining the integrity of scene semantics. Distinct from DLPL, PPTFormer utilizes a more lightweight contourlet transformation for texture extraction, followed by keypoint-based perspective description. To further reduce network complexity, we employ a non-parametric method for perspective prototype construction based on linear moving averages, coupled with a simplified Gaussian mixture distribution for generating new pseudo perspectives efficiently. We also utilize a more efficient all-MLP visual decoder instead of the complex architecture in DLPL.  Moreover, to address domain shift in the multi-perspective learning process, a challenge which DLPL has overlooked, we improve the sequential attention mechanism (PIA in DLPL) with the iterative Pseudo Multi-Perspective (PMP) Attention, as well as utilize a lightweight Perspective Calibration process following the PPTFormer Blocks.

Our contributions can be summarized as follows:
\begin{itemize}
   
\item  We propose a lightweight PPTFormer that enables implicit multi-perspective learning in the absence of multi-perspective datasets.  By generating pseudo multi-perspective characterization about the scene and engaging in joint learning across them, PPTFormer can effectively simulate the varying viewpoints encountered during actual UAV flight, thereby improving the segmentation accuracy.

\item  Particularly, we employ contourlet transformation and keypoint-based perspective description, coupled with non-parametric perspective prototype construction. Moreover, a perspective calibration process is utilized to address domain shift in multi-perspective learning.

\item  PPTFormer achieves competitive performance across five UAV segmentation datasets while maintaining a lightweight network, demonstrating its effectiveness in capturing the intricate dynamics of UAV-captured scenes through pseudo multi-perspective Learning.

\end{itemize}

\begin{figure*}
    \centering
    \includegraphics[width=1\linewidth]{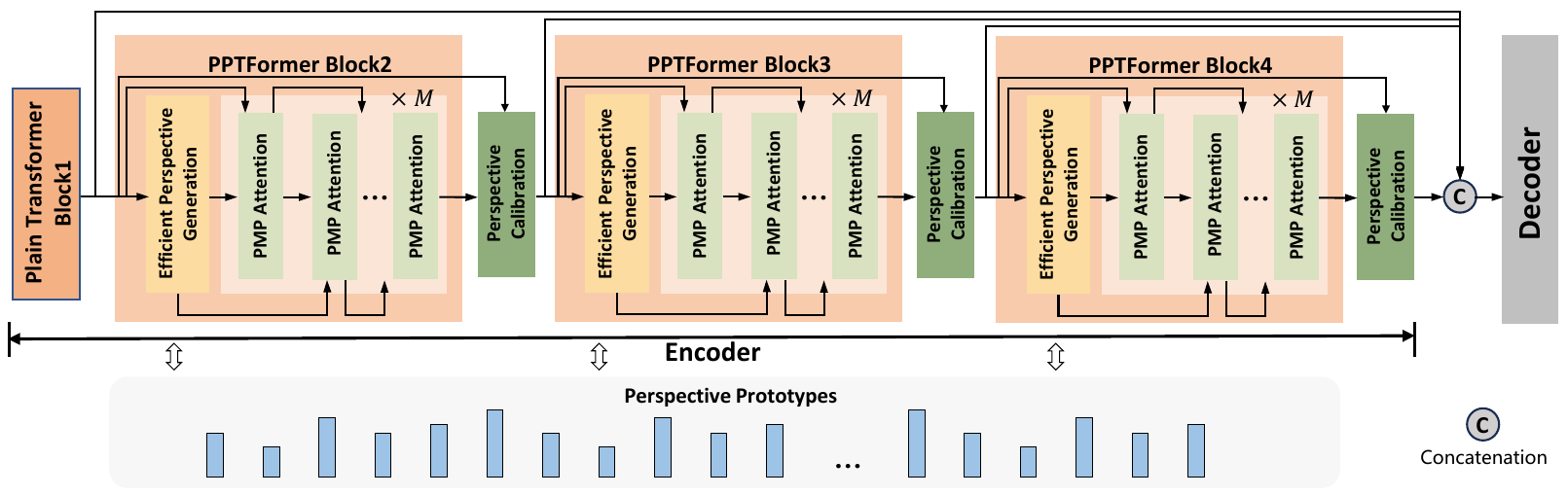}
    \caption{The overview of the proposed PPTFormer. 
    The encoder comprises four Transformer Blocks: one Plain Transformer Block followed by three PPTFormer Blocks. 
    The former is responsible for extracting basic low-level information, which serves as the foundation for pseudo multi-perspective learning in the subsequent blocks.
    Interleaved between the PPTFormer Blocks are Perspective Calibration modules, which prevent potential scene domain shifts. Finally, we concatenate features of varying scales produced by each block and feed them into the decoder network.
    }
    \label{fig_method}
\end{figure*}

\section{Related Work}

\subsection{Semantic Segmentation}

Semantic Segmentation has been a classical and fundamental task in computer vision area \cite{chen2024reason3D,homview,cagcn,zhu2024ibd,wang2021learning,ipgn,feng2018challenges,mscnn,changenet,zhu2024fouridown,zhu2023exploring,zhu2023learning,yu2023learning}.
In recent years, the majority of segmentation advancements are grounded in the use of fully convolutional networks (FCN) \cite{zhu2023gabor,zhu2023continual,fcn,zheng2022unsupervised,zheng2023learning,yu2022frequency,NEURIPS2023_043f0503,yu2023learning}.
Subsequent research has concentrated on capturing contextual relationships within images, employing sophisticated network architectures to enhance the understanding of scene composition \cite{cdgc,sstkd,zhu2023llafs}.

\subsection{UAV Scene Segmentation}

Despite these advancements, there is a noticeable gap in the literature pertaining to semantic segmentation tailored for UAV imagery. Existing UAV segmentation methods mainly focus on solving the class imbalanced problems, such as SCO \cite{sco}, FarSeg \cite{farseg} and  PointFlow \cite{pointflow}. 
The unique challenges posed by UAV scenes, characterized by diverse and dynamic changes in perspective, are seldom addressed. The development of segmentation models that can effectively handle the variability inherent in UAV-captured images also remains an area in need of further exploration.

\subsection{Ultra-High Resolution Segmentation}

In comparison to natural scene imagery, images captured from Unmanned Aerial Vehicles (UAVs) generally exhibit higher resolution characteristics. Existing research has also attempted to introduce segmentation algorithms that can concurrently balance accuracy and efficiency, such as WSDNet \cite{urur}, GPWFormer \cite{gpwformer}, among others. In this paper, we endeavor to enhance segmentation precision from the perspective of UAV flight viewpoints. This approach is universally applicable and can be integrated with the aforementioned high-resolution image segmentation algorithms.

\section{PPTFormer}
\label{sec_method}

\subsection{Overall Structure}

The overall architecture of our proposed Pseudo Multi-Perspective Transformer (PPTFormer) is inspired by \cite{dlpl} and  depicted in Figure \ref{fig_method}. Following the classic paradigms \cite{setr,segformer}, PPTFormer consists of a meticulously designed encoder for perspective learning and a generic decoder. The encoder comprises four Transformer Blocks: one Plain Transformer Block followed by three PPTFormer Blocks. The former is responsible for extracting basic low-level information, which serves as the foundation for pseudo multi-perspective learning in the subsequent blocks. Unlike \cite{dlpl} which applies DLPL at every Transformer Block, we restrict perspective learning to only the last three Blocks in PPTFormer to enhance its efficiency. Specifically, within the PPTFormer Blocks, we extract implicit perspective representations from the visual features based on the structural texture. In conjunction with training across the entire dataset, we create perspective prototypes of the images present throughout the dataset. These prototypes are shared across the three PPTFormer Blocks to ensure a consistent learning of perspectives during the training process. Interleaved between the PPTFormer Blocks are Perspective Calibration modules, which are instrumental in aligning the visually fused features from pseudo perspectives with the original perspective of the image. This alignment prevents potential scene domain shifts, which are overlooked in DLPL \cite{dlpl}. Finally, we concatenate features of varying scales produced by each block and feed them into the decoder network for further processing.

\subsection{PPTFormer Block} \label{sec_pptblock}

As shown in Figure \ref{fig_method}, the PPTFormer Block  comprises an Efficient Perspective Generation module and $M$ layers of Pseudo Multi-Perspective Attention (PMP Attention). Given the input of low-level visual features $F$ from block 1, the Efficient Perspective Generation module implicitly represents  the image's perspective $p$, generating a pseudo perspective $p'$ that simulates the movement and shift of viewpoints during an actual UAV flight, all while preserving the semantic information of the scene. During this process, the acquired perspective representation $p$ contributes to the construction of perspective prototypes $P$ for the entire dataset and also bases the pseudo perspective generation on these prototypes. The output visual feature $F'$ with the pseudo perspective $p'$, along with $F$, are both fed into the $M$ layers of PMP Attention for multi-perspective correlation. This allows the model to understand the scene's semantic information from both the original perspective and the new pseudo perspective simultaneously. Specifically, in the first layer of PMP Attention, the inputs are $F$ and $F'$, and the output is the first level of perspective fusion. Subsequently, in the following $M-1$ layers of PMP Attention, the fused feature and $F'$ work together to achieve the subsequent $M-1$ levels of perspective fusion. Below, we will introduce the specific structures of the Efficient Perspective Generation and PMP Attention in detail.

\begin{figure*}
    \centering
    \includegraphics[width=1\linewidth]{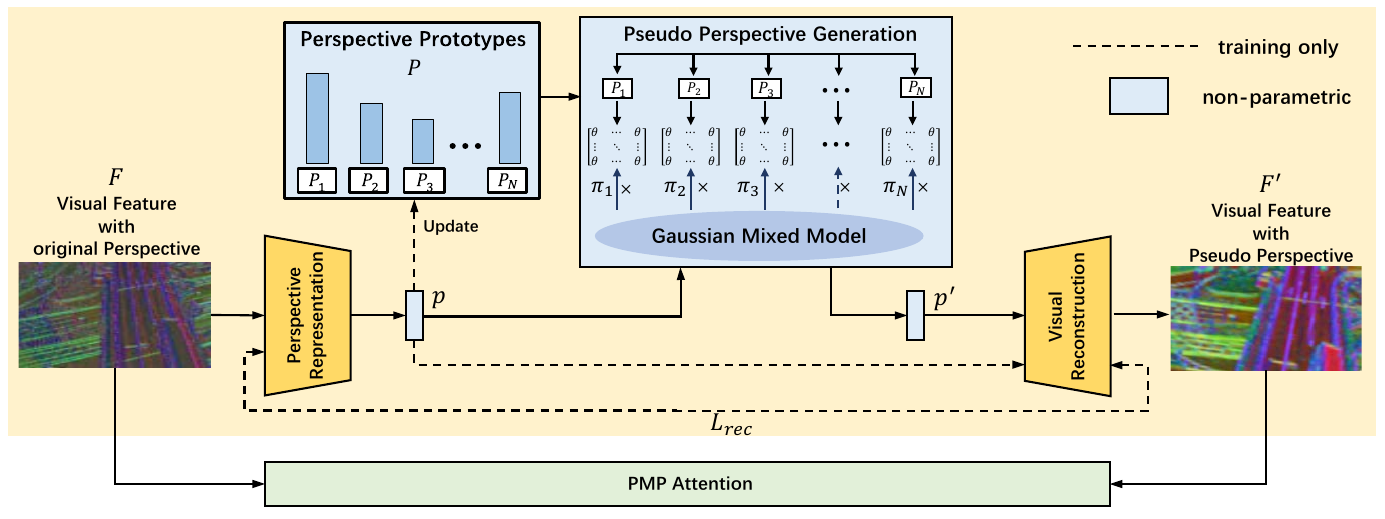}
    \caption{The Efficient Perspective Generation module in PPTFormer Block.
    }
    \label{fig_pt}
\end{figure*}

\subsection{Efficient Perspective Generation}

As illustrated in Figure \ref{fig_pt}, the input is the visual feature $F$ form Block 1, which first passes through a Perspective Representation encoder $\mathbf{E}_{p}$ to obtain an original Perspective $p$. Subsequently, on one hand, $p$ contributes to the construction of the entire dataset's perspective prototypes $P$ using the online sequential clustering updating technique. The length of $P$, which corresponds to the number of perspective prototypes in the dataset, is $N$. On the other hand, $p$ is also used for Pseudo Perspective Generation based on $P$, to generate a new pseudo perspective $p'$. Thereafter, a Visual Reconstruction decoder $\mathbf{D}_{p}$ uses $p'$ to ultimately reconstruct the visual feature $F'$.  In 
order to reduce the complexity,  $\mathbf{D}_{p}$ is all-MLP architecture which demonstrates promising performance here. During training, to ensure its reconstructive capability, $p$ is also directly fed into $\mathbf{D}_{p}$ with the aim of restoring the original visual feature $F$, that is,
\begin{equation}
    L_{rec} = ||\mathbf{D}_p(p) - F||_2 = ||\mathbf{D}_p(\mathbf{E}_p(F)) - F||_2,
\end{equation}
\noindent where $L_{rec}$ is the reconstruction loss.

\subsubsection{Perspective Representation}

\begin{figure}
    \centering
    \includegraphics[width=0.88\linewidth]{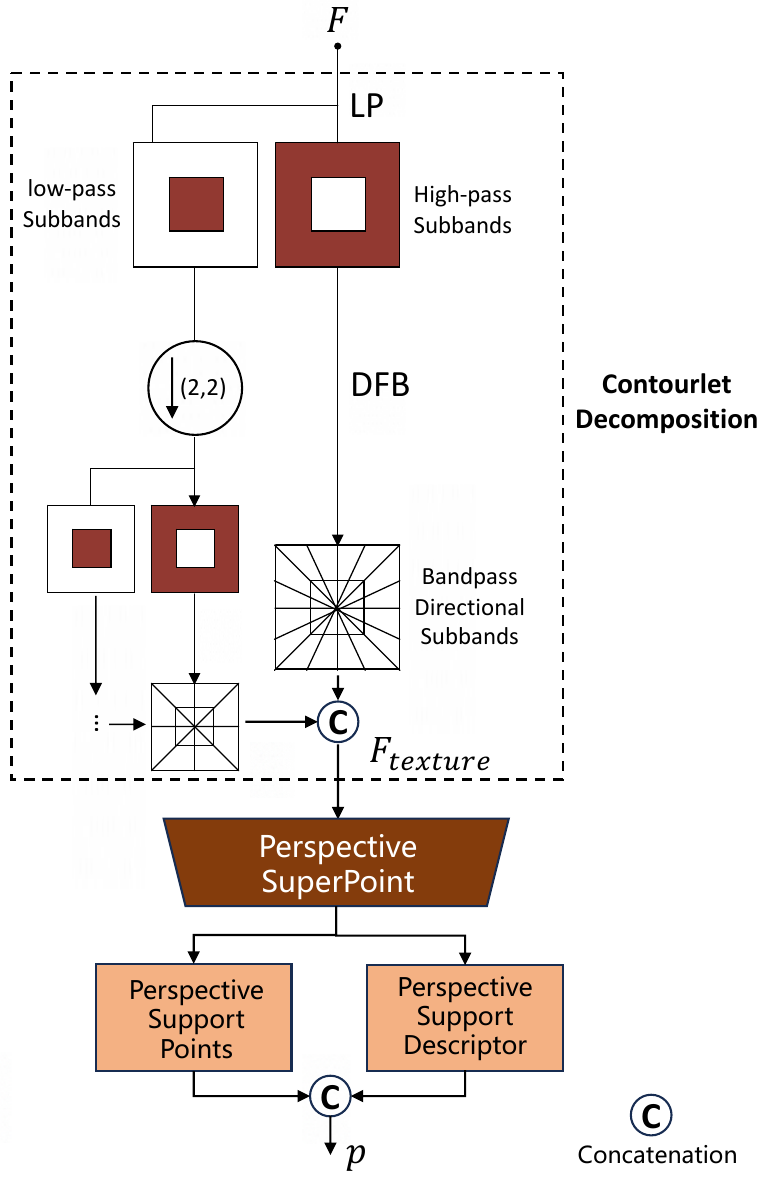}
    \caption{The detailed process of Perspective Representation.}
    \label{fig_pr}
\end{figure}

As depicted in Figure \ref{fig_pr}, distinct from DLPL, PPTFormer utilizes a more lightweight contourlet transformation for texture extraction, followed by keypoint-based perspective description. The Perspective Representation encoder $\mathbf{E}_{p}$ primarily encompasses two processes: extracting low-level structural texture from the image using contourlet decomposition \cite{contourlet} and, based on this texture, extracting interest super points that are related to the image's perspective. The former ensures that the model captures the global structural texture information, which can represent the image's contours, edges, and other structural features, including perspective information. To further distill perspective-related features, the latter identifies key support points representing perspective as super points, whose spatial distribution and feature intensity can construct a structured high-dimensional description of the image's perspective.

\paragraph{Texture Extraction.} Specifically, as traditional filters, contourlet decompositions inherently excel at texture representation across various geometric scales and directions in the spectral domain. Rather than describing texture features in the spatial domain, they analyze the energy distribution in the spectral domain to extract the inherent geometric structures of the texture, which naturally includes the image's perspective.

The contourlet decomposition comprises a cascaded Laplacian Pyramid (LP) \cite{lp} and a directional filter bank (DFB) \cite{dfb}. The LP decomposes input features into low-pass and high-pass subbands using pyramidal filters.
The high-pass subband is processed through the DFB, which is employed to reconstruct the original signal with minimal sample representation, produced by $t$-level binary tree decomposition in the two-dimensional frequency domain, resulting in \( 2^z \) directional subbands. For instance, when \( z = 3 \), the frequency domain is divided into 8 directional subbands, with subbands 0-3 and 4-7 corresponding to vertical and horizontal details, respectively  \cite{sstkd}. Following \cite{sstkd}, for a richer expression, we stack multiple contourlet decomposition layers iteratively, and concatenate the output of each level to form the final extracted structural texture. 

Specifically, the output of level $t\in [1, T]$ is denoted as $F_{bds, t}$, 
\begin{equation}
    \begin{aligned}
        F_{bds, t} &= {\rm \mathbf{DFB}}(F_{h, t}), ~~~ t \in [1, T], \\
        & {\rm where} ~~   F_{l, t}, F_{h, t}={\rm \mathbf{LP}}(F_{l, t-1}) \\
    \end{aligned}
    \label{eq_con}
\end{equation}
\noindent where $l$ and $h$ represent the low-pass and high-pass subbands respectively, $bds$ denotes the bandpass directional subbands. Then structural texture $F_{texture}$ is denoted as:
\begin{equation}
    F_{texture} = \mathop{\rm Cat}\limits_{t \in [1, T]} \{F_{bds, t}\}.
\end{equation}

\noindent where ${\rm Cat}$ is the concatenation operation. $F_{texture}$ is rich in texture information including the image's perspective. 

\paragraph{Perspective Support Description.} Based on \( F_{texture} \), we use a SuperPoint network to extract key support points and corresponding support descriptors that are capable of characterizing the perspective from the texture features. Specifically, this network comprises two parallel heads, $\mathbf{S}_{sp}$ and $\mathbf{S}_{sd}$, which respectively output the ``point-ness" probability map and the corresponding point feature descriptor. The final output, the perspective feature \( p \), is the concatenation of output features from the two heads, along the channel dimension:
\begin{equation}
    p = {\rm Cat}(\mathbf{S}_{sp}(F_{texture}), \mathbf{S}_{sd}(F_{texture})).
\end{equation}

\subsubsection{Perspective Prototypes Construction}

The construction of Perspective Prototypes is aimed to obtain and manage the scene perspective types of the whole dataset, by performing an online sequential clustering process on the coming $p$s. We utilize a lightweight memory bank and its length $N$ is equal to the number of prototypes. Specifically, different from DLPL \cite{dlpl}, we employ a non-parametric mechanism to reduce the network complexity.  Firstly, the $N$ prototypes $P = \{P_1, P_2, ..., P_n, ..., P_N\}$ are initialized with the first input $N$ $p$s, and we set the counts $\{c_1, c_2, ..., c_n, ..., c_N\}$ to record the number of perspective features belonging to the corresponding prototype. Then, for each new coming $p$, we find its closest prototype $P_n$ by L2 distances, and update the the prototype with:
\begin{equation}
\begin{aligned}
    c_n &\leftarrow c_n + 1 \\
    P_n &\leftarrow P_n + \frac{1}{c_n} (p - P_n). 
\end{aligned}
\end{equation}
\noindent So the final resulting prototype $P_n$ is the moving average of $p$s that are closest to $P_n$.

Then we formulate overall perspective distribution of the whole dataset in form of Gaussian Mixed Model (GMM), following DLPL,
\begin{equation}
    G(P) = \sum_{n=1}^{N} \pi_n \cdot  \mathcal{N}(p | P_n, \Sigma_n),
\end{equation}
\noindent where $\mathcal{N}(\cdot)$ indicates the Gaussian Distribution, the $n$th component of GMM has the center of $P_n$ with the variance of $\Sigma_n$, and $\pi_n$ is the mixture coefficient of meets: 

\begin{equation}
\begin{aligned}
    \sum_{n=1}^N \pi_n = 1, ~~~0 \le \pi_n \le 1,
\end{aligned}
\end{equation}

\subsubsection{Pseudo Perspective Generation }

Based on the dynamically updated perspective distribution $G(P)$, we can generate a new semantic-related pseudo perspective $p'$ of the given probe $p$, by leveraging all the prototypes. Note that we avoid using Homography transformations in DLPL to accelerate the generating process.

\begin{equation}
\begin{aligned}
    p' = p \cdot G(P, p)  = p \cdot \sum_{n=1}^{N} \pi_n \cdot  \mathcal{N}(p | P_n, \Sigma_n) ,
\end{aligned}    
\end{equation}

\noindent where $p'$ is generated based on the overall perspective distribution over all perspective prototypes.

Finally, the corresponding visual feature $F'$ for $p'$ can be reconstructed with $\mathbf{D}_{p}$.
\begin{equation}
    F' = \mathbf{D}_{p}(p').
\end{equation}

\subsection{Pseudo Multi-Perspective Attention}

By the Efficient Perspective Generation, we obtain the visual feature $F'$ with a new pseudo perspective for the input image. As seen that $F$ and $F'$ contain closely identical scene context and structured information but only differ in perspective. Next, they are fed into the $M$ layers of Pseudo Multi-Perspective (PMP) Attention to leverage the relationship between the $F$ (with original perspective $p$) and $F'$ (with generated pseudo perspective $p'$). Different from DLPL which use sequential attention mechanism, the PMP Attention employs an iterative one,
\begin{equation}
\begin{aligned}
    {\rm PMP\_Att}&(F, F') = \\ 
    &\left\{
                                \begin{aligned}
                                 &F_1 = {\rm Softmax}(\frac{F \times F^{'\top}}{\sqrt{C}}) \times F',   \\ 
                                 &F_2 = {\rm Softmax}(\frac{F' \times F_{1}^{\top}}{\sqrt{C}}) \times F_1, \\
                                 &F_3 = {\rm Softmax}(\frac{F_1 \times F_{2}^{\top}}{\sqrt{C}}) \times F_2, \\
                                 & ... \\
                                 &F_M = {\rm Softmax}(\frac{F_{M-2} \times F_{M-1}^{\top}}{\sqrt{C}}) \times F_{M-1}.
                                \end{aligned}
                                \right.,
\end{aligned}
\end{equation}

\noindent where $C$ is the feature channel. By the iterative attention calculation, the multiple perspectives can be fully correlated.

\subsection{Perspective Calibration}

Within each PPTFormer Block, after undergoing $N$ layers of PMP Attention, the original perspective and the pseudo perspective are thoroughly fused multiple times, ultimately enabling the model to capture scene information as observed from various perspectives. However, in practice, we observed that as the perspective fusion progresses, there could be some domain shift within the visual features' depiction of the scene. To prevent such occurrences, we further incorporate a straightforward Perspective Calibration process after PPTFormer Blocks. Specifically, this entails passing the visual feature with the original perspective, which is the input to the current PPTFormer Block, through a skip connection to calibrate the fused feature output by the current PPTFormer Block, by several layers of PMP Attention. In practice, we found this elegant approach to be effective in mitigating issues of domain shift.

\subsection{Optimization}

The overall loss function $L$ is the combination of the main segmentation loss $L_{seg}$  and the reconstruction loss $L_{rec}$:
\begin{equation}
    L = L_{seg} + \lambda L_{rec},
    \label{loss}
\end{equation}
\noindent where $\lambda$ is the weight for $L_{rec}$, and set to 0.4.

\section{Experiments}

\subsection{Datasets and Evaluation Metrics}

In our experiments, we validate the effectiveness of PPTFormer on five datasets, including UDD6, iSAID, UAVid , Aeroscapes, and DroneSeg.

\subsubsection{UDD6} Urban Drone Dataset (UDD) dataset is collected by a DJI-Phantom 4 UAV at altitudes between 60 and 100 meters, and is extracted from 10 video sequences. The resolution is either 4k (4096$\times$2160) or 12M (4000$\times$3000).
It contains a variety of urban scenes. 

\subsubsection{iSAID} 
iSAID totally consists of 2,806 images, where 1411, 458, and 937 images are for training, validation, and testing sets, respectively.

\subsubsection{UAVid} UAVid dataset has 300 images of size of 3840$\times$2160, where the training, validation, and testing set contains 200, 70, and 30 images respectively.

\subsubsection{Aeroscapes} 
The Aeroscapes  dataset provides 3,269 720p images and ground-truth masks for 11 categories, where the training and validation sets include 2,621 and 648 images respectively.

\subsubsection{DroneSeg} The DroneSeg dataset \cite{dlpl} extends the segmentation annotations from VisDrone  dataset. The dataset consists of 10,209 images with fine-grained pixel-level annotations of 14 categories.

\subsection{Implementation Details}

In our experiments, we  adopt the MMSegmentation  toolbox as codebase and follow the default augments without bells and whistles.  SuperPoint network is used for the perspective support description.
To ensure training stability, during the initial 30\% of epochs, we replace the PMP Attention with plain Self-Attention. This substitution aims to guarantee the reliability of perspective representation and reconstruction within $\mathbf{E}_p$ and $\mathbf{D}_p$, as well as the stability of learning perspective prototypes. Subsequently, we revert to the PMP Attention mechanism to perform global joint optimization in the remaining epochs.
In the training, SGD optimizer with momentum 0.98 for all parameters is used, the initial learning rate is configured as  5 $\times 10^{-3}$ and the maximum iteration number is set to 160K for all datasets. In Eq. \ref{eq_con}, $T$ is set to 2. The length of $P$ is set to $N=64$.

\begin{table}[]
\centering
\scalebox{0.83}{\begin{tabular}{c|ccccc}
\toprule
\multirow{2}{*}{\textbf{Method}}         & \multicolumn{5}{c}{\textbf{mIoU} (\%)}                                    \\ \cmidrule{2-6} 
                                                              & \textbf{UDD6}  & \textbf{iSAID} & \textbf{UAVid} & \textbf{Aeroscapes} & \textbf{DroneSeg} \\ \midrule
Deeplab        & 71.84 & 59.20                    & 56.82 & 51.40      & 38.69    \\
OCR\_W48                       & 73.37 & 62.73                    & 63.10 & 58.19      & 43.10    \\
PSPNet                                               & 72.95 & 60.30                    & 58.20  & 57.98      & 37.03    \\
FarSeg                                           & -     & 63.70                    & -     & -          & -        \\
FarSeg++                 & -     & 63.70                    & -     & -          & -        \\
PFNet                                                  & -     & 66.90                    & -     & -          & -        \\
SCO                                        & -     & 69.10                    & -     & -          & -        \\ \midrule
SETR                                                        & 68.00 & 62.77                    & 58.52 & 50.34      & 48.23    \\
UperNet                                                    & 73.13 & 66.45                    & 61.91 & 64.32      & 53.34    \\
PoolFormer                                                & 74.54 & 65.55                    & 61.73 & 62.27      & 53.94    \\
SegFormer                                                  & 74.28 & 67.19                    & 62.01 & 66.40      & 55.33    \\ \midrule
\textbf{PPTFormer}  & \textbf{76.70} & \textbf{69.87}                    & \textbf{65.00} & \textbf{68.50}      & \textbf{57.71}   \\ \bottomrule
\end{tabular}}
\caption{Comparison with state-of-the-arts on UDD6, iSAID, UAVid, Aeroscapes, and our proposed DroneSeg datasets.}
\label{exp_sota}
\end{table}

\subsection{Comparison with State-of-the-Arts}

We compare PPTFormer with both representative CNN-based (FarSeg \cite{farseg}, FarSeg++ \cite{farseg}, PFNet \cite{pointflow} and SCO \cite{sco}) and ViT-based (SETR \cite{setr}, UperNet \cite{swin}, PoolFormer \cite{poolformer}, SegFormer \cite{segformer}) segmentation methods on five benchmark datasets. 

\subsubsection{UDD6, UAVid}
 
Both the two datasets contain relative fewer images and lower scene complexity and we compare their results here. For fair comparisons with ViT-based methods, we adopt large backbones for CNN-based methods including ResNet-101 and HRNet-W48. As shown in Table \ref{exp_sota}, the ordinary transformer (SETR) show even lower performance than CNN-based methods, and the advanced ones including PoolFormer, SegFormer shows better results. The proposed PPTFormer achieves further performance improvements on both the datasets.

\subsubsection{iSAID, Aeroscapes, DroneSeg}

These three datasets consist of more images than UDD6 and UAVid, and have higher scene perspective variances. So they would be more convincing to prove the superiority. PPTFormer outperforms other methods by a larger margin, which demonstrates the effectiveness of the proposed method on the description of perspective information.

\subsection{Ablation Study}

All ablation studies are performed on DroneSeg \textit{testing} set, SegFormer is used as baseline network.

\begin{table}[]
\centering
\scalebox{1}{\begin{tabular}{c|c}
\toprule
Perspective-Oriented Learning Method &  mIoU (\%) \\ \midrule
Baseline (SegFormer w/o data aug.) & 52.03  \\ \midrule
+ Random Rotate & 52.94    \\
+ Random Scale  & 53.09    \\
+ Random Perspective-Vertical   & 53.56  \\
+ Random Perspective-Horizontal   & 53.42   \\ 
+ Random Combination & 55.33    \\  \midrule
\textbf{PPTFormer}  & 57.71    \\\bottomrule
\end{tabular}}
\caption{The comparison of PPTFormer  with other perspective-oriented learning methods (data augmentation). Perspective-Vertical and Perspective-Horizontal means adjusting perspectives in vertical and horizontal directions, and are implemented by the default ``torchvision.transforms" interfaces in PyTorch.}
\label{exp_compare_view}
\end{table}

\subsubsection{Comparison with Perspective Learning Methods}

Given that PPTFormer is a perspective-oriented learning approach, we begin by comparing it with various perspective-based augmentations. We observe that in UAV scenarios, perspective shifts are almost invariably linked to changes in altitude and angular positioning, which manifest as alterations in scale and rotation. Therefore, we employ a combination of these two data augmentation techniques to benchmark against PPTFormer. As illustrated in Table \ref{exp_compare_view}, we discover that utilizing either augmentation method in isolation yields only modest enhancements over the baseline approach. In contrast,  PPTFormer secures a substantial increase in performance. We further demonstrate that PPTFormer remains compatible with standard data augmentations for additional gains.

 \begin{table}[]
\centering
\scalebox{1}{\begin{tabular}{c|c}
\toprule
Contourlet Decomposition  &  mIoU (\%)  \\ \midrule
0 & 56.88   \\ 
1 & 57.40   \\ 
2 & 57.71   \\ 
3 & 57.73   \\ \bottomrule
\end{tabular}}
\caption{The impact of contourlet decomposition.}
\label{exp_co}
\end{table}

\begin{table}[]
\centering
\begin{tabular}{c|c}
\toprule
Method  & Parameters \\ \midrule
SegFormer      & 84.7M     \\
DLPL(Segformer) & 87.5M     \\
PPTFormer       & 86.0M     \\ \bottomrule
\end{tabular}
\caption{Efficiency analysis.}
\label{table_param}
\end{table}

\subsubsection{The Impact of Contourlet Decomposition}

The contourlet decomposition is capable of extracting structural texture information from images, which encompasses a wealth of perspective details. By initially employing it within the Perspective Representation, the network can swiftly focus on shallow image textures, thereby facilitating further extraction of Perspective and enhancing learning efficiency. Table \ref{exp_co} demonstrates its efficacy, as the number of contourlet decomposition layers increases, the mIoU correspondingly improves. Here, a layer count of zero indicates no application of contourlet decomposition.

\subsubsection{Effectiveness of Perspective Calibration}

\begin{figure}
\centering
  \includegraphics[width=0.9\linewidth]{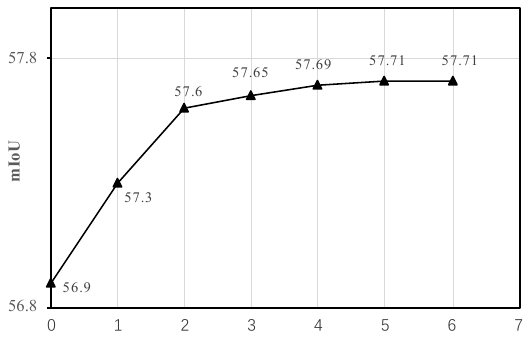}
  \caption{Layer number in perspective calibration.}
  \label{exp_center_num2}
\end{figure}

The purpose of Perspective Calibration is to prevent the occurrence of scene domain shift that may arise as a consequence of deep perspective fusion. Figure \ref{exp_center_num2}  illustrates the impact of the number of PMP Attention layers in it on model performance. ``Layer=0" implies the absence of Calibration, and the results indicate a low mIoU under this condition. As the number of layers increases, there is a significant improvement in mIoU, which underscores the effectiveness and necessity of Perspective Calibration.

\subsubsection{The Quantity of Perspective Prototypes}
\begin{figure}
\centering
  \includegraphics[width=0.9\linewidth]{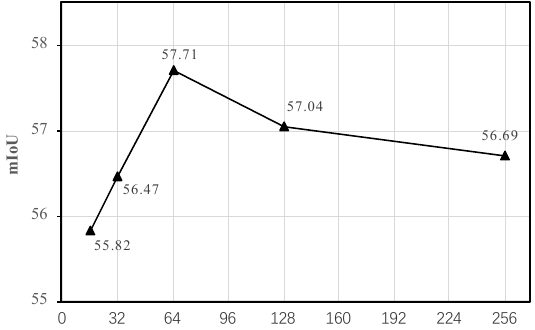}
  \caption{The quantity of perspective prototypes.}
  \label{exp_center_num}
\end{figure}

The quantity of perspective prototypes represents the entirety of perspective variations found within the dataset, with a higher count enabling the retention of a more extensive set of prototypes.  Figure \ref{exp_center_num}  reveals that with a smaller allocation of prototypes (16, 32), the process fails to exhaustively capture all perspectives, resulting in an underfitting of the model. Conversely, as the number of prototypes increases (128, 256), we generate an overly dense array of prototypes. This surplus can introduce redundancy and give rise to numerous discrete perspectives, potentially hindering the learning process.

\subsubsection{Efficiency Analysis}

PPTFormer, inspired by the theoretical framework of DLPL, exhibits a more lightweight architecture and enhanced efficiency. In Table \ref{table_param}, we present a comparative analysis of the parameter counts between PPTFormer and DLPL (based on SegFormer). The comparison indicates that PPTFormer introduces only a slight increase in parameters over the baseline SegFormer, yet maintains a lower parameter count than DLPL. This demonstrates that PPTFormer not only achieves high accuracy but also sustains a high level of efficiency.

\section{Conclusion}

This paper presents the novel PPTFormer, a Pseudo Multi-Perspective Transformer network for UAV scene segmentation. It addresses the challenges of capturing the dynamic perspectives inherent in UAV-captured imagery. By integrating systematic Pseudo Multi-Perspective Learning within the Transformer framework, PPTFormer adeptly performs efficient pseudo perspective generation and achieves multi-perspective correlation, leading to a more nuanced understanding of UAV scenes. The experiments on several datasets validate the superior performance and efficiency of PPTFormer. The significant advancements made by PPTFormer underscore the importance of perspective-oriented learning in semantic segmentation and pave the way for further innovation in the processing of UAV-captured visual data.

\section*{Acknowledgments}
This work was supported by the Anhui Provincial Natural Science Foundation under Grant 2108085UD12, 
the National Key R\&D Program of China under Grant 2020AAA0103902, NSFC (No. 62176155), Shanghai Municipal Science and Technology Major Project, China (2021SHZDZX0102). We acknowledge the support of GPU cluster built by MCC Lab of Information Science and Technology Institution, USTC.

\section*{Contribution Statement}

The first two authors contribute equally to this work.

\bibliographystyle{named}
\bibliography{ijcai24}

\end{document}